\documentclass[sigconf]{acmart}

\AtBeginDocument{%
  \providecommand\BibTeX{{%
    \normalfont B\kern-0.5em{\scshape i\kern-0.25em b}\kern-0.8em\TeX}}}
\setcopyright{acmcopyright}
\copyrightyear{2022}
\acmYear{2022}
\acmDOI{XXXXXXX.XXXXXXX}

\acmConference[ACM SIGKDD '22]{ACM SIGKDD Workshops}{August 14--18, 2022}{Washington, DC}

\acmBooktitle{ACM SIGKDD '22: ACM SIGKDD Workshops,
 August 14--18, 2022, Washington, DC} 
\acmPrice{15.00}
\acmISBN{978-1-4503-XXXX-X/22/08}

\usepackage{multirow}
\usepackage{algorithm}
\usepackage{algpseudocode}

\begin{document}

\title{Multiple Instance Learning for Detecting Anomalies over Sequential Real-World Datasets}

\author{Parastoo Kamranfar}
\affiliation{%
  \institution{George Mason University}
  \streetaddress{4400 University Drive}
  \city{Fairfax}
  \state{Virginia}
  \country{USA}
  \postcode{22030}}
  \email{pkamranf@gmu.edu}

\author{David Lattanzi}
\affiliation{%
 \institution{George Mason University}
 \streetaddress{4400 University Drive}
 \city{Fairfax}
 \state{Virginia}
 \country{USA}
 \postcode{22030}}
 \email{dlattanz@gmu.edu}

\author{Amarda Shehu}
\affiliation{%
  \institution{George Mason University}
  \streetaddress{4400 University Drive}
  \city{Fairfax}
  \state{Virginia}
  \country{USA}
  \postcode{22030}}
  \email{ashehu@gmu.edu}
  
\author{Daniel Barbar\'{a}}
\authornote{Corresponding author.}
\affiliation{%
  \institution{George Mason University}
  \streetaddress{4400 University Drive}
  \city{Fairfax}
  \state{Virginia}
  \country{USA}
  \postcode{22030}}
  \email{dbarbara@gmu.edu}

\renewcommand{\shortauthors}{Kamranfar, et al.}

\begin{abstract}
Detecting anomalies over real-world datasets remains a challenging task. Data annotation is an intensive human labor problem, particularly in sequential datasets, where the start and end time of anomalies are not known. As a result, data collected from sequential real-world processes can be largely unlabeled or contain inaccurate labels. These characteristics challenge the application of anomaly detection techniques based on supervised learning. In contrast, Multiple Instance Learning (MIL) has been shown effective on problems with incomplete knowledge of labels in the training dataset, mainly due to the notion of bags. While largely under-leveraged for anomaly detection, MIL provides an appealing formulation for anomaly detection over real-world datasets, and it is the primary contribution of this paper. In this paper, we propose an MIL-based formulation and various algorithmic instantiations of this framework based on different design decisions for key components of the framework. We evaluate the resulting algorithms over four datasets that capture different physical processes along different modalities. The experimental evaluation draws out several observations. The MIL-based formulation performs no worse than single instance learning on easy to moderate datasets and outperforms single-instance learning on more challenging datasets. Altogether, the results show that the framework generalizes well over diverse datasets resulting from different real-world application domains.
\end{abstract}



\maketitle
\section{Introduction}
\label{sec:Introduction}

Anomaly detection (AD) is a well-studied problem in machine learning and seeks to detect data points/instances that deviate from an expected behavior; the deviating instances are known as outliers or anomalies~\cite{chandola2009anomaly}. 
Applications of AD have been explored extensively in literature and over a variety of datasets, including data generated from sequential real-world processes~\cite{blazquez2021review,chandola2008comparative}. 

Despite being well-studied, AD over real-world datasets remains challenging. In many real-world settings, annotation of sequential data requires intensive human labor; annotation is also made difficult by unknown or noisy start and end times of anomalies. As a result, data collected from sequential real-world processes are often largely unlabeled, or labels are often inaccurate. This setting is particularly challenging for AD methods formulated under the umbrella of supervised learning.

In contrast, Multiple Instance Learning (MIL) has been shown effective in dealing with incomplete knowledge of labels in the training dataset. MIL was introduced in~\cite{dietterich1997solving} in the context of binary classification as a weakly-supervised approach that reduces annotation efforts. MIL assumes that the data points/instances are organized in sets, also known as bags. Essentially, MIL deals with bag labels instead of individual instance labels~\cite{guan2016efficient}. The label of a bag  (negative/positive) is assigned according to the instance labels the bag contains. The existence of at least one positive instance is enough for labelling the bag as positive, while negative bags need to include all negative instances~\cite{feng2021multiple,carbonneau2018multiple}. This provides a way of building a more robust classifier than what can be built by relying exclusively on single-instance labels.

MIL has received much attention in the machine learning community in the past two decades, and various techniques have been developed to exploit the structure of the data to enhance performance in a variety of applications~\cite{quellec2016multiple, zhang2007multiple, andrews2002support, AlamShehuBICOB20, zhang2021multiple, AlamShehuJBCB21}. In particular, computer vision, where videos are naturally structured into bags and require significant single-instance labeling efforts, has benefited tremendously from MIL-based methods. Application of MIL for AD remains largely under-leveraged~\cite{amores2013multiple,foulds2010review,yang2013TRASMIL}. Yet, MIL provides an appealing formulation for semi-supervised AD over real-world datasets, and it is the primary contribution of this paper. 

In this paper, we propose an MIL-based formulation and propose various algorithmic instantiations of this framework based on different design decisions for key components of the framework. In particular, we leverage the combination of MIL and the Strangeness based OUtlier Detection (StrOUD) algorithm~\cite{barbara2006detecting}. StrOUD computes a strangeness/anomaly value for each data point and detects outliers by means of statistical testing and calculation of p-value. Thus, the framework is dependent on two primary design decisions: definition of the strangeness factor and the aggregation function. The degree of outlying/strangeness is needed for recognizing anomalous data points, while the aggregate function is needed to aggregate the measures of strangeness into a single anomaly score of a bag. In this paper, we utilize two alternative scores, the 'Local Outlier Factor' and the Autoencoder (AE) reconstruction error. We utilize six aggregate functions (minimum, maximum, average, median, spread, dspread) for the strangeness measure of a bag. We note that we are not limited to these design decisions, and they can be easily generalized to any other choices. To the best of our knowledge, the effort we describe here is the first to truly leverage the power of MIL in AD by hybridizing the concept of bags naturally with AD methods that are routinely used in the field.

We evaluate various algorithmic instantiations of the MIL framework over four datasets that capture different physical processes along different modalities (video and vibration signals). The results show that the MIL-based formulation performs no worse than single instance learning on easy to moderate datasets and outperforms single-instance learning on more challenging datasets. Altogether, the results show that the framework generalizes well over diverse benchmark datasets resulting from different real-world application domains.

The rest of the paper is organized as follows. Section~\ref{sec:PriorWork} relates prior work in AD. The proposed methodology is described in Section~\ref{sec:Methods}, and the experimental evaluation is related in Section~\ref{sec:Results}. Section~\ref{sec:Conclusion} concludes the paper.
\section{Prior Work}
\label{sec:PriorWork}

Extensive studies have been performed on AD in many application domains, from fraud detection in credit cards, to structural health monitoring in engineering, to bioinformatics in molecular biology~\cite{kou2004survey,zhang2021anomaly, AlamShehuJBCB21}. Different types of data have been considered, from sequential time series data~\cite{blazquez2021review,chandola2009detecting}, to image data, to molecular structure data~\cite{da2020critical,zenati2018adversarially, AlamShehuJBCB21}.

Many methods have been developed for AD, varying from traditional density-based methods~\cite{ide2007computing,tang2002enhancing} to more recent AE-based ones~\cite{chen2018autoencoder,babaei2021aegr}. Density-based methods assign an anomaly score to a single data point/instance by comparing the local neighborhood of a point to the local neighborhoods of its $k$ nearest neighbours. Higher scores are indicative of anomalous instances.

The Local Outlier Factor (LOF) and its variants are density-based anomaly scores that are utilized extensively in AD literature~\cite{hautamaki2004outlier,ahmed2017anomaly}. For instance, work in~\cite{tang2002enhancing} introduced an LOF variant score called 'Connectivity based Outlier Factor' (COF) which differs from LOF in the way that the neighborhood of an instance is computed. 'Outlier Detection using In-degree Number' (ODIN) score has been presented in~\cite{hautamaki2004outlier}. ODIN measures the number $k$ of nearest neighbors of a data point which also have that data point in their neighborhood. The inverse of ODIN is defined as the anomaly score.  

There are other methods, known as deviation-based methods, that also utilize anomaly scores. These methods attempt to find a lower-dimensional space of normal data by capturing the correlation among the features. The data are projected onto a latent, lower-dimensional subspace, and unseen test data points with large reconstruction errors are determined to be anomalies. As these methods only encounter normal data in the training phase that seeks to learn the latent space, they are  known as semi-supervised methods.

PCA- and AE-based methods are in the category of AD methods that seek to capture linear and non-linear feature correlations, respectively~\cite{shyu2003novel,chen2018autoencoder}. Due to the capability of AEs in finding more complex, non-linear correlations, AE-based AD methods tends to perform better than PCA-based ones, generating fewer false anomalies. 

In ~\cite{chen2018autoencoder}, conventional and convolutional AE-based (CAE) methods are presented and compared with PCA-based methods. Work in~\cite{an2015variational} proposes a variational AE-based (VAE) method which takes advantage of the probabilistic nature of VAE. The method leverages the reconstruction probability instead of the reconstruction error as the anomaly score. AE-based methods, however, require setting a threshold for how large the reconstruction error or reconstruction probability has to be for an instance to be predicted as anomalous.

Generally, most AD methods require careful setting of many parameters, including the anomaly threshold, which is an ad-hoc process. This hyperparameter regulates sensitivity to anomalies and the false alarm rate (rate of normal instances detected as outliers/anomalies), and is the indicator of AD performance. 
To maintain a low false alarm rate, conformal AD (CAD) methods build on the conformal prediction (CFD) concept~\cite{laxhammar2011sequential}.      

The underlying idea in CFD methods is to predict potential labels for each test data point by means of the p-value (one p-value per possible label). The non-conformity measure is utilized as an anomaly score, and p-values are calculated  
To this end, the significance level needs to be determined in order to retain or reject the null hypothesis~\cite{laxhammar2015inductive}.

In their utilization of significance testing to avoid overfitting, CFD methods are highly similar to the classic Strangeness based OUtlier Detection (StrOUD) method~\cite{barbara2006detecting}. StrOUD was proposed as an AD method that combines the ideas of transduction and hypothesis testing. It eliminates the need for anomaly ad-hoc thresholds and can additionally be used for dataset cleaning. 

Apart from determination of the anomaly score or the AD category, there are many fundamental challenges in AD, not the least of which is finding appropriate training instances. It is generally easier to obtain instances from the normal behaviour of a system rather than anomalies, especially in real-world settings (i.e. large engineering infrastructures or industrial systems)~\cite{villa2021semi}, where anomalous physical processes that generate anomalous data are rare events.

As a result, there is strong potential for weakly-supervised techniques such as MIL for AD. MIL has been thoroughly explored for classification problems in a variety of applications, such as face detection~\cite{zhang2007multiple}, text categorization~\cite{andrews2002support}, and speech context classification~\cite{zhang2021multiple}, but it has been largely under-explored for AD. A local AD method, TRASMIL, has been proposed in~\cite{yang2013TRASMIL} for abnormal event detection in video clips. The method was constructed on two main concepts: trajectory segmentation and MIL. Trajectories are defined as bags, and they are partitioned into sub-trajectories to form the instances.
Another connection between MIL and AD in machine learning literature is work in~\cite{quellec2016multiple}, which employs MIL on medical images for the purpose of beast cancer diagnosis in mammography screening. In this work, each image is considered as a group of regions, and features are extracted from each region. A bag is composed of regions from four images, and labels are assigned to mammography examinations instead of individual images. Recent works in video surveillance~\cite{sultani2018real,feng2021mist} employ MIL to perform anomaly detection. However, they do so in the context of weakly supervised anomaly detection, assuming some anomalous labels are available. Our approach is completely unsupervised, since we wish to apply it to scenarios where such labels are not available.    

In this paper we leverage MIL due to its appealing formulation for semi-supervised AD over real-world datasets. Furthermore, to avoid the determination of an anomaly threshold, our MIL framework builds on the StrOUD method. The LOF and AE reconstruction error are considered  as alternative anomaly scores. Various aggregation functions are considered, and in turn various algorithmic instantiations are obtained, effectively resulting in an ablation study. Methodological details now follow.
\section{Methodology} 
\label{sec:Methods}

As we pointed out in Section \ref{sec:Introduction}, the proposed framework adapts the ideas of MIL and StrOUD to detect anomalous test bags. We first summarize key ingredients of MIL and StrOUD to set the stage for how we hybridize them in a novel MIL-StrOUD framework.

StrOUD was initially presented in~\cite{barbara2006detecting} for the purpose of finding outliers in a pre-defined clustering setting. It utilizes the concept of transduction to directly make estimates for each data point within the training/reference set and so reason from these cases to the test set. In essence, the StrOUD algorithm places an unknown data point in a known data distribution (the baseline), and determines the fitness of the point by carrying out statistical testing. 

In evidential studies, a null hypothesis is formulated first, and then a statistical testing is conducted to retain or reject the null hypothesis. The p-value as the outcome of such statistical testing is a probability value used as a measurement of evidence against the null hypothesis. The smaller the p-value, the stronger the evidence to reject the null hypothesis~\cite{du2009confidence}. In statistics, by making the assumption that the null hypothesis is correct, the p-value is defined as the probability of observing results that can be considered at least as extreme as the observed results from statistical testing. To this end, StrOUD takes advantage of a strangeness value for every point of interest and utilizes the p-value significance testing formulation to support or reject a null hypothesis (i.e. a point $i$ belongs to baseline/normal distribution).  

MIL on the other hand assumes that data points are organized in sets called bags~\cite{carbonneau2018multiple}. MIL can be categorized as weakly-supervised binary classification; instead of dealing with labels of individual instances within a bag, it works with negative or positive labels that are assigned to each bag~\cite{feng2021multiple,carbonneau2018multiple}. This provides a more robust classifier than the one which is built by relying exclusively on instance labels and relaxes the dependency on individual instance diagnosis. A classic MIL assumption is that a negative bag only contains negative instances (normal), while a positive bag contains at least one positive instance (anomalous) known as a witness. The bag classifier, $f(X)$, is defined as: 

\begin{equation}
  f(X)=\begin{cases}
    1, & if \exists{b \in B: g(x) = 1} \\
    0, & \text{otherwise}.
  \end{cases}
\end{equation}

where $b$ and $B$ denotes an instance and a bag, respectively. $g: X \rightarrow \{0, 1\}$ is a mapping process that corresponds negative and positive instances to the $0$ and $1$ classes~\cite{carbonneau2018multiple}. Due to the goal of prediction at the bag-level, there is no need to identify all witnesses in a bag, and existence of one suffices to label a bag as positive (label of '1').

The MIL-StrOUD approach we propose in this paper is shown in pseudocode in Algorithms~\ref{alg:create-baseline} to~\ref{alg:bag-score}. It uses the same strategy as MIL classification or AE-based anomaly detection, where data needs to be split into training and testing sets. We note that in the MIL setting training includes only normal data, while the testing set contains both normal data and anomalies.

There is no constraint on the number of bags or the number of instances inside a bag, but we know that bag composition (the number of instances within a bag) has an impact on the performance of MIL-based approaches~\cite{carbonneau2018multiple}. There is no standard/general solution on how to organize bags in MIL literature. Bag size is typically determined empirically for a problem/data at hand~\cite{guan2016efficient}. In our problem, by reasoning on data and attempting to achieve a balanced number of positive and negative bags, we define the number of instances in bags for each dataset. This process is explained in section~\ref{sec:Results}.

\begin{algorithm}
{\footnotesize
\begin{algorithmic}[1]
\caption{Create Baseline Anomaly Distribution}
\label{alg:create-baseline}
\Function{CreateBaseline}{training bags ${rb_1, rb_2,...,rb_k}$}

\For{$\forall rb_i \in X_{train}$}
\State {$AnomalyScore_{rb_i}\gets FindAnomalyScore({rb_i})$}
\State {$AnomalyScore_{X_{train}} \gets Append(AnomalyScore_{rb_i})$}
\EndFor 

\State {$OrderedAnomalyList_{X_{train}} \gets Sort(AnomalyScore_{X_{train}}, ascending)$}

\Return{$OrderedAnomalyList_{X_{train}}$}

\EndFunction
\end{algorithmic}}
\end{algorithm}

\begin{algorithm}
{\footnotesize
\begin{algorithmic}[1]
\caption{Find Anomaly of a Query Bag}
\label{alg:query-bag}
\Function{FindQueryBagAnomaly}{query bag ${qb}$, baseline anomaly ${OrderedAnomalyList_{X_{train}}}$}
\State $prediction_{qb} \gets 0$

\State {$AnomalyScore_{qb} \gets FindAnomalyScore({qb})$}

\State {$index_{qb} \gets $index of $AnomalyScore_{qb}$ in $OrderedAnomalyList_{X_{train}}$}

\State $p-value_{qb} \gets \frac{1+n_x-index_{qb}}{|n_x|+1}$

\For{$\forall c_i \in ConfidenceRange$}
\State $\tau \gets 1 - c_i$
    \If {$p_{qb} <= \tau$}
        \State $prediction_{qb} \gets 1$
    \EndIf
    \State $Compare (Ground Truth_{qb}, Prediction_{qb})$
\EndFor \\
\Return{}

\EndFunction
\end{algorithmic}}
\end{algorithm}

\begin{algorithm}
{\footnotesize
\begin{algorithmic}[1]
\caption{Find Bag Anomaly Score}
\label{alg:bag-score}
\Function{FindAnomalyScore}{bag ${b}$}

\For{$\forall x_i \in b$}
\State $AnomalyScore_{x_i} \gets LOF(b, x_i)$ or $MSE (b,x_i)$
\State $AnomalyScoreList_{b} \gets Append(AnomalyScore_{x_i})$
\EndFor 
\State $AnomalyScore_{b} \gets Aggregate(AnomalyScoreList_{b}, Function)$
 
\Return{$AnomalyScore_{b}$}

\EndFunction
\end{algorithmic}
}
\end{algorithm}

Algorithm~\ref{alg:create-baseline} creates a baseline anomaly distribution from training bags which all are normal. For this purpose, the anomaly score of each bag needs to be computed first. This is what algorithm~\ref{alg:bag-score} carries out. Once the values are computed, all anomaly scores of the reference bags (${rb_1, rb_2,...,rb_k}$) are merged, and a list containing strangeness values of training set (${AnomalyScore_{X_{train}}}$) is then created (line $2$--$5$). The final step is to order these anomaly measurements and return the baseline list which is needed for the p-value computation of a query/test bag.  

Algorithm~\ref{alg:query-bag} predicts the label of a query bag ($qb$) by means of the baseline strangeness distribution calculated in Algorithm~\ref{alg:create-baseline}. The function starts with the assumption that $qb$ is 'normal' (line $2$). Like the reference bags, the anomaly score of $qb$ should be computed by calling the function in~\ref{alg:bag-score}. Once the strangeness of the bag is computed through the aggregate function, it can be compared to a sample distribution of normal bags (baseline distribution) whose strangeness is computed using the same aggregate function.

To predict the label of $qb$, we utilize the hypothesis testing part of the StrOUD algorithm. Thus, we need to find the appropriate index of $qb$, where it should be inserted to maintain the order of the baseline (line $4$). The next step is to determine the $p-value$ of the query bag by considering a fraction of reference bags whose anomaly scores are equal or greater than that of $qb$ (line $5$). This probability value is used for statistical analysis, where the null hypothesis is that the $qb$ belongs to normal distribution. 

Therefore, the next block (lines $6$--$12$) is designed to support or reject the null hypothesis by defining a confidence level $c_{i}$. If the $p-value$ is less than the confidence threshold, we reject the null hypothesis and the bag is predicted as 'anomalous.' The prediction evaluation is done by comparing with the ground truth (line $11$). 

Algorithm~\ref{alg:query-bag} is given for a single test bag but needs to be repeated for all testing bags. To better evaluate the results, we report the area under the receiver-operating-characteristic (ROC) curve (AUC) instead of accuracy~\cite{guan2016efficient}. A wide range of confidence levels (from $0$ to $1$ with $0.001$ step) are examined, and AUC values are calculated by means of $False Positive Rate$ and $True Positive Rate$. 
   
Many MIL-based algorithms used for classification of bags solve the problem by predicting the class of each participating instance in a bag as a side feature. We follow a similar procedure by first computing the degree of outlying/strangeness for each individual instance inside a bag.
Algorithm~\ref{alg:bag-score} is designed to do so. As related earlier, we consider two different approaches to compute the strangeness factor for an instance, LOF or MSE (lines $2$--$3$). LOF or MSE constitute the first important design decision of our proposed MIL-StrouD framework.

The strangeness factor determination is the first design decision of the framework we propose. From all these strangeness scores, a single measure of a bag should be obtained (line $6$). We propose the use of an aggregate function $f_{a}(\bar{s})$, where the $\bar{s}$ is the vector of strangeness of instances, namely ($s_{1},s_{2},...,s_{k}$) and $k$ is the number of instances in a bag. There are many choices for $f_{a}(\bar{s})$, including, but not limited to, the maximum, minimum, average, median, spread, and dspread functions. The aggregation function is the second design decision of our proposed MIL-StrOUD framework. 

Details of the design choices are elaborated as follows.

\subsection{I. Strangeness Factor}

\subsubsection{Local Outlier Factor (LOF)} 

LOF is known as a density-based anomaly detection factor that captures relative degree of isolation for each data point. In summary, LOF detects local outliers with respect to their neighborhood and with the assumption that outliers are far away from inliners located in a dense region of the neighborhood. This value is computed by determination of a \emph{MinPts} hyperparameter identifying the expected number of neighbors around each data point/instance~\cite{breunig2000lof}. A low LOF score is an indication of an inliner, while a high value shows an anomalous point. After computing the proximity matrix from all training/reference and test points to all reference points, as explained in~\cite{babaei2021aegr}, the LOF calculation process is performed in four steps.
    
First, the k-nearest distances (\emph{k-distance}) for all reference and test points to every reference point are computed. 

Second, the reachability distance (RD) is then calculated as:
\vspace{-3mm}

\begin{equation}
    RD (a,b) = max\{k-distance(b), Euclidean-dist(a,b)\}
\end{equation}
    
which depends on the query point location in the neighborhood, the output is either \emph{k-distance} or the actual Euclidean distance between two points. 
    
In the third step, the Local Reachability Distance (LRD) needs to be calculated. LRD can be interpreted as an estimated distance at which a point can be found by its neighbors. This value is computed by inverting the average reachability distance of a point from its neighbors   
\vspace{-3mm}
    
\begin{equation}
LRD (a) = \frac{N_{MinPts(a)}}{\sum_{b\in{N_{MinPtns(a)}}} RD(a,b)}
\end{equation}
    
where $N_{MinPts}$ is the LOF hyperparameter which is the number of neighbors in data point neighborhood.
    
In the fourth and last step, the LRD ratio of the point and its neighbors are measured to determine if the point is close to its neighbors to be labelled as an inlier or is far away to be labelled as anomaly. 
\vspace{-3mm}
    
\begin{equation}
LOF (a) = \frac{\sum_{b\in N_{MinPts(a)}} \frac{LRD(b)}{LRD (a)}}{N_{MinPts(a)}}
\end{equation}

\subsubsection{Autoencoder (AE) Mean Reconstruction Error (MSE)} 

An AE is a neural network composed of two parts, an encoder and decoder. The encoder maps an input ($x$) to the latent space ($h$), while the decoder reconstructs the original input from the hidden representation by the same transformation used in encoder ($z$)~\cite{tschannen2018recent}. The reconstruction error defines as the difference between the original $x$ and reconstructed $z$, and an AE learns to minimize this error in training phase. An encoder and decoder equations of a single hidden layer AE are given in~\ref{eq:autoencoder}~\cite{an2015variational}. 

\begin{equation}
\label{eq:autoencoder}
h = {g(W_{e}x+b_{e})} \qquad  z = {g(W_{d}x+b_{d})}
\end{equation}
    
where $g$ is a nonlinear transformation function, $W_{e}$ and $W_{d}$ are encoder and decoder weights, while $b_{e}$ and $b_{d}$ are network biases. 
    
We employ the same idea as AE-based anomaly detection which utilizes the reconstruction error as strangeness factor. The underlying assumption is that, as the AE is trained only on normal data, normal points in the test set ought to be reconstructed well but not anomalous data. The higher the reconstruction error, the higher an anomaly score of a point.  

We empirically select the AE architectures for each dataset. In all four architectures, three dense hidden layers along with two dropouts are utilized. An $80-20$ split is followed to split the training dataset into training and validation sets. We use the \emph{tanh} activation function in the first hidden layer and two \emph{ReLu}s in the other two hidden layers. 
In general, the number of neural units are gradually dropped in each layer until the desired latent dimensions are reached. Different loss functions are also examined for each dataset. Furthermore, the number of training epochs is determined by plotting training versus validation loss over the number of epochs. We have explored many variations of this fundamental architecture, varying the number of hidden layer, the layer width (number of neurons), the activation functions, and the number of latent dimensions (data not shown). We adopt the notation $x \rightarrow L1 \rightarrow L2 \rightarrow y$ to denote the encoder architecture, with the decoder being a mirror image. We have experimented with $|L_1| \in [100, 512]$, $|L_2| \in [25, 128]$, and $|y| \in [5, 32]$.

\subsection{II. Aggregate Function}

We define below the six different aggregate functions we utilize for different instantiations of our framework:
    
    \begin{itemize}
      \item The maximum strangeness, that is $f_{a}(\bar{s})=max(s_{1},s_{2},...,s_{k})$
      \item The minimum strangeness, that is $f_{a}(\bar{s})=min(s_{1},s_{2},...,s_{k})$
      \item The average strangeness, that is $f_{a}(\bar{s})=mean(s_{1},s_{2},...,s_{k})$
      \item The median strangeness, that is $f_{a}(\bar{s})=median(s_{1},s_{2},...,s_{k})$
      \item The dspread strangeness, that is $f_{a}(\bar{s})=mean(s_{1},s_{2},...,s_{k})+2 * stdev(s_{1},s_{2},...,s_{k})$
      \item The spread strangeness, that is $f_{a}(\bar{s})=mean(s_{1},s_{2},...,s_{k}) * stdev(s_{1},s_{2},...,s_{k})$
\end{itemize}

In all the above, $s_1, s_2, \ldots, s_k$ are samples/instances in a bag $\bar{s}$. We allow for these various aggregation functions in our experimental evaluation, but we do expect that the maximum strangeness will confer better performance. The reason for this is tied to the definition of an anomalous bag. As described above, we adopt the definition of a bag as anomalous as long as there is at least one anomalous sample in it. It is worth noting, however, that this definition is conventionally employed in applications of MIL for classification (even though this restriction is relaxed in cases where positive bags cannot be identified by a single instance but by the accumulation of several of them~\cite{carbonneau2018multiple}), and we adopt it here for MIL for AD, as a first step. Future research can consider alternative definitions.

\section{Results}
\label{sec:Results}

We first provide a brief summary of each of the four datasets before relating the experimental evaluation.

\subsection{Datasets}

\emph{I. Case Western Reserve University (CWRU)}: The CWRU bearing dataset is a standardized dataset for pattern classification with mechanical systems. It has known ground-truth labels associated with five motor-bearing condition states (one 'normal' versus four 'damaged'). More details about this dataset can be found in~\cite{yang2018bearing}. All four damaged states' measurements are considered as anomalies. The data contains $1853$ instances; $300$ normal instances are utilized for the training dataset, while the rest constitute the testing dataset. We divide the data into $30$ normal bags and $35$ anomalous bags in order to have a balanced number of normal and anomalous test bags. 

\emph{II. Smartphone-Based Recognition of Human Activities and Postural Transitions Data Set (HAPT):} HAPT is a publicly-labelled dataset obtained from basic static, dynamic activities, and postural transitions performed by $30$ volunteers, captured by a smartphone, and pre-processed by applying filters and using fixed-length sliding window with overlap. Each data instance is a $561$-featured vector containing time and frequency domain variables. More details can be found in the UCI repository~\cite{Dua:2019}. Here, we define postural transition instances ($518/10,929$ of the data instances) as anomalous activities. To balance the number of bags, $104$ training bags are extracted from $5,200$ normal instances (static and dynamic activities,) while $104$ test bags are formed from the remaining $5,211$ data points for test set; $518$ outliers are randomly assigned to half of the test bags to generate $52$ normal and anomalous bags. Each bag contains at least $50$ instances.

\emph{III. Virat Video Dataset:} Virat is a well-known, public video surveillance dataset. It contains data collected from multiple outdoor scenes showing people doing normal activities. Typical uses of the dataset are for activity recognition; more details can be found in~\cite{oh2011large}. We consider one out of $16$ available scenes. There are multiple video clips from a scene, and each clip can contain from zero to multiple activities. Normal instances are defined as those with normal human activities. Video clips that show humans with objects, such as scooters or bicycles are defined as anomalies. In this setting, the video clips are natural bags; we note that in this dataset, there are no single-instance labels but only clip/bag labels. Our training and test datasets contain $2,000$ images each, organized into $24$ and $34$ bags, respectively. We utilize the pre-trained InceptionV3 model on ImageNet~\cite{szegedy2016rethinking} for feature extraction. By adding an extra layer of \emph{GlobalMaxPooling}, we obtain $2048$ features to encode each image/instance.    

\emph{IV. Bridge Ambient Vibration Dataset}
This dataset was collected from the Old ADA bridge in Japan over four different bridge states (ground-truth labels) of the bridge: 1) INT: no damage; 2) DMG1: half cut of one vertical member truss at the bridge midspan; 3) DMG2: fully cut of the same member; 4) DMG3: fully cut of one vertical member at the bridge 5/8th span. The experiment was conducted at 200 Hz sampling rate. Depending on a damage state, the test was repeated for a varying number of times. More details about the experiment can be found in~\cite{kim2021ambient}. Signals are segmented into $1$ second of measurements to generate the instances. 
We have included all three INT test repetitions in our analysis; the first and third are considered for training, and the second is reserved for testing. $30$ seconds of measurements from the normal and $60$ seconds of measurements from the three damaged states are considered in our bag composition, leading to $622$ ($21$ training bags) and $1,433$ data instances in the training and test sets ($14$ normal vs. $17$ anomalous testing bags), respectively.

\subsection{Experimental Results}
We relate our experiments on each of the four datasets in order. In each, we compare the AUC obtained by MIL versus StrOUD; as described in Section~\ref{sec:Methods}, the latter is a natural choice for single-instance learning, given our leverage of the LOF in one instantiation of MIL. We consolidate the comparisons of the MIL-derived algorithms to compare the incorporation of LOF versus MSE (by autoencoder). In the interest of space, we suppress the varying instantiations based on the aggregation function, and only note when presented results are representative or in contrast indicative of superior performance due to any aggregation function in particular. We note that the number of nearest neighbors needs to be specified when using LOF, but the aggregate function is only needed for MIL due to its bag-level prediction. For all datasets, the same aggregate functions are employed, but the number of neighbors depends on the number of instances in a bag. So, this value is related for each dataset separately. When MSE is considered, the best autoencoder architecture and model are identified empirically for each dataset. 

\subsubsection{Comparative Evaluation on the CWRU Dataset}

CWRU is considered a clean dataset. We employ it here as a baseline or control setting. Due to the existence of at least $10$ instances in each bag, a range of $2$ to $10$ number of nearest neighbors is examined for the LOF computation in the MIL-StrOUD setting; in StrOUD, where there is no such a limitation, a larger range of values from $2$ to $50$ is considered. In the encoder architecture employed for MIL-StrOUD with MSE, $|x| = 400$, $|L_1| = 200$, $|L_2| = 50$, and $|y| = 10$.

The AUC comparison is related in Table~\ref{tab:CWRUResults}. It is evident that no matter what anomaly score is employed, both algorithms achieve the highest AUC; we do not repeat the results but instead relate LOF/MSE in Table~\ref{tab:CWRUResults}. The results confirm that anomalies behave differently enough from normal data in the CWRU dataset that they can be distinguished even with single-instance learning. For LOF, the results achieved with the \emph{maximum} and \emph{dspread} aggregates are the same, while for MSE, \emph{maximum} and \emph{spread} aggregates achieved the top performance.

\begin{table}[htbp]
\centering
\caption{Results on the CWRU Dataset.}
\label{tab:CWRUResults}
\vspace*{-5mm}
\begin{tabular}{|l|l|l|}
\hline
Method & Strangeness Factor & Best AUC \\\hline
\multirow{1}{*}{MIL-StrOUD} & LOF/MSE & $1.0$ \\
                            \hline
\multirow{1}{*}{StrOUD}     & LOF/MSE & $1.0$ \\
                           \hline
\end{tabular}
\end{table}
\vspace*{-3mm}

\subsubsection{Comparative Evaluation on the HAPT Dataset}

We consider a similar range of number of neighbors for the MIL setting on the HAPT dataset ($2$ to $10$), but a larger range of neighbours for StrOUD ($2$ to $70$), as the HAPT dataset contains more instances than the CWRU dataset. We note that in the encoder architecture employed for MIL-StrOUD with MSE, $|x| = 561$, $|L_2| = 280$, $|L_3| = 35$, and $|y| = 5$. 

\begin{figure}[htbp]
\begin{tabular}{c}
\textbf{MIL-StrOUD with LOF}\\
\includegraphics[width=0.43\textwidth]{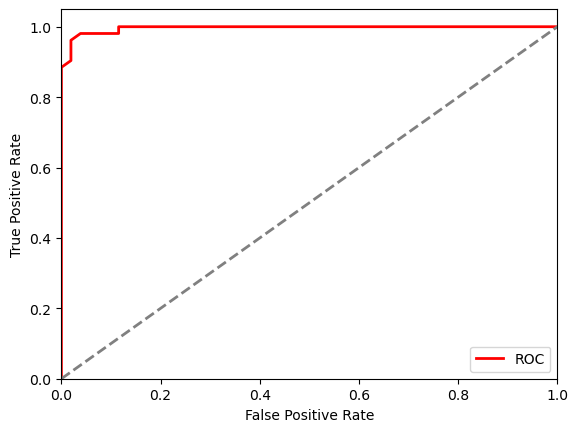}\\[1mm]
\textbf{MIL-StrOUD with MSE}\\
\includegraphics[width=0.43\textwidth]{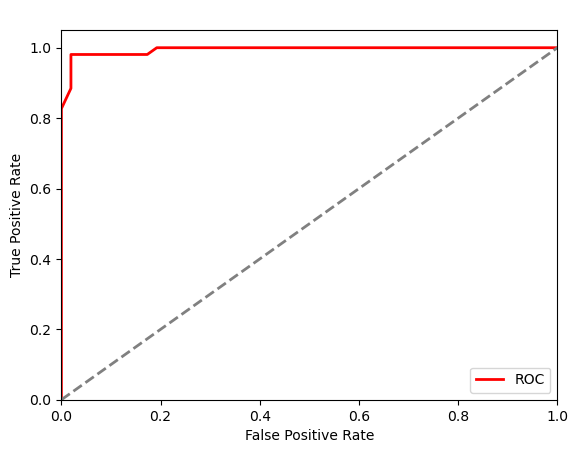}\\
\end{tabular}
\vspace*{-7.5mm}
\caption{ROC curves are related for MIl-StrOUD with LOF (top panel) and MSE (bottom panel) for the HAPT dataset. These correspond to the setting that confers the highest performance, related in Table~\ref{tab:HAPTResults}.}
\label{fig:HAPT-ROCs}
\end{figure}

Table~\ref{tab:HAPTResults} relates the comparative evaluation. As HAPT is a clean benchmark dataset, the AUCs of the MIL-based algorithms and StrOUD are close. This indicates that the extracted features are meaningful enough to distinguish postural transitions from other activities, and that labels are accurately assigned to the instances. We do observe MIL being able to produce slightly better AUCs than StrOUD using either LOF or MSE score. Fig.~\ref{fig:HAPT-ROCs}, which shows the ROC curves for MIL-StrOUD with LOF (top panel) or MSE (bottom panel) additionally relates the high performance of MIL for this dataset.

\begin{table}[htbp]
\centering
\caption{Results on the HAPT Dataset.}
\label{tab:HAPTResults}
\vspace*{-5mm}
\begin{tabular}{|l|l|l|}
\hline
Method & Strangeness Factor & Best AUC \\\hline
\multirow{2}{*}{MIL-StrOUD} & LOF & $0.996$ \\
                            & MSE & $0.994$ \\ \hline
\multirow{2}{*}{StrOUD}     & LOF & $0.990$ \\
                            & MSE & $0.980$ \\ \hline
\end{tabular}
\end{table}

\vspace{-2mm}

In general, we observe that the results related above are  not overly sensitive to the number of nearest neighbors; that is, the AUC varies in a narrow range from $0.964$ to $0.996$, achieving the best value for  $NN = 2 $. The \emph{maximum} aggregate function is the one conferring the highest performance to MIL, demonstrating that the assumption of a single anomaly in a bag is sufficient for determining the bag to be anomalous holds for this dataset.


\subsubsection{Comparative Evaluation on the Virat Dataset}
\begin{figure}[htbp]
\begin{tabular}{c}
\textbf{MIL-StrOUD with LOF}\\
\includegraphics[width=0.43\textwidth]{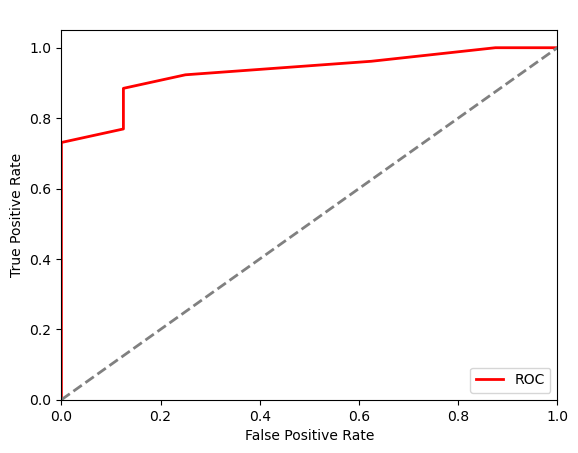}\\[1mm]
\textbf{MIL-StrOUD with MSE}\\
\includegraphics[width=0.43\textwidth]{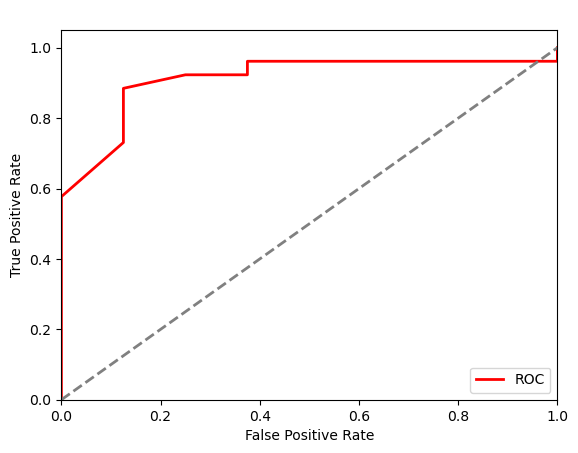}\\
\end{tabular}
\vspace*{-7.5mm}
\caption{ROC curves are related for MIl-StrOUD with LOF (top panel) and MSE (bottom panel) for the VIRAT dataset. These correspond to the setting that confers the highest performance, related in Table~\ref{tab:ViratResults}.}
\label{fig:VIRAT-ROCs}
\end{figure}

We assign bag labels manually based on the objects of the scene in the VIRAT dataset. However, instance-level labeling cannot be carried out for this dataset, as such information is not available, and the process is too laborious. The performance of the MIL algorithms on the Virat dataset is related in Table~\ref{tab:ViratResults}. The comparison shows higher performance by LOF over MSE. Fig.~\ref{fig:VIRAT-ROCs}, which shows the  ROC curves for MIL-StrOUD with LOF (top panel) or MSE (bottom panel) supports these results. 

We note that in the encoder architecture employed for MIL-StrOUD with MSE, $|x| = 2048$, $|L_2| = 512$, $|L_3| = 128$, and $|y| = 32$. We observe that the results are mildly sensitive to the number of nearest neighbors; that is, there is a wide range of the number of nearest neighbors where the performance ranges from $0.897$ to $0.937$, reaching the latter at $NN =3$. We again observe the \emph{maximum} aggregate function as the one conferring the highest performance to MIL, as in the previous dataset.

\begin{table}[htbp]
\centering
\caption{Results on the Virat Dataset.}
\vspace*{-5mm}
\label{tab:ViratResults}
\begin{tabular}{|l|l|l|}
\hline
Method & Strangeness Factor & Best AUC \\\hline
\multirow{2}{*}{MIL-StrOUD} & LOF & $0.937$ \\
                            & MSE & $0.911$ \\ \hline
\end{tabular}
\end{table}
\vspace{-2mm}

\vspace*{-3mm}

\subsubsection{Comparative Evaluation on the Bridge Dataset}
According to the minimum number of instances in each bag ($30$), a range of $2$ to $20$ neighbours is examined for MIL. This value ranges from $2$ to $50$ for StrOUD. The encoder architecture employed for MIL-StrOUD with MSE, $|x| = 200$, $|L_2| = 100$, $|L_3| = 25$, and $|y| = 5$.

Table~\ref{tab:BridgeResults} relates the comparison of MIL-based algorithms to StrOUD. It is clearly evident that using MIL improves the results significantly over StrOUD. Fig.~\ref{fig:VIRAT-ROCs} relates the  ROC curves for MIL-StrOUD with LOF (top panel) or MSE (bottom panel). We observe that the results on the Bridge dataset are moderately sensitive to the number of nearest neighbors; that is, for a wide range of values, the AUC ranges from $0.895$ to $0.931$ at $NN=2$. Again, the \emph{maximum} aggregate function is the best choice.

\begin{table}[htbp]
\centering
\caption{Results on the Bridge Dataset.}
\label{tab:BridgeResults}
\vspace*{-5mm}
\begin{tabular}{|l|l|l|}
\hline
Method & Strangeness Factor & Best AUC \\\hline
\multirow{2}{*}{MIL-StrOUD} & LOF & $0.931$ \\
                            & MSE & $0.905$ \\ \hline
\multirow{2}{*}{StrOUD}     & LOF & $0.720$ \\
                            & MSE & $0.729$ \\ \hline
\end{tabular}
\end{table}
\vspace*{-3mm}

\begin{figure}[htbp]
\begin{tabular}{c}
\textbf{MIL-StrOUD with LOF}\\[-1mm]
\includegraphics[width=0.43\textwidth]{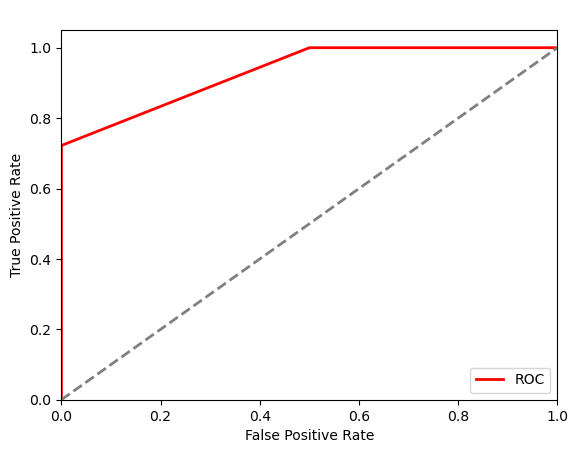}\\[0.5mm]
\textbf{MIL-StrOUD with MSE}\\[-1mm]
\includegraphics[width=0.43\textwidth]{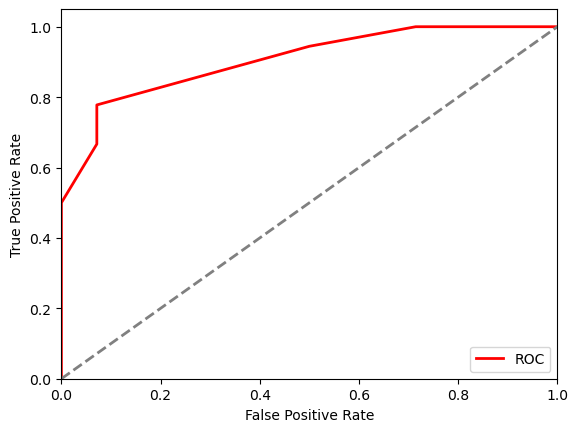}\\
\end{tabular}
\vspace*{-7.5mm}
\caption{ROC curves are related for MIl-StrOUD with LOF (top panel) and MSE (bottom panel) for the Bridge dataset. These correspond to the setting that confers the highest performance, related in Table~\ref{tab:BridgeResults}.}
\label{fig:BRIDGE-ROCs}
\end{figure}

\begin{figure}
\centering
\begin{tabular}{cc}
\includegraphics[width=0.43\textwidth]{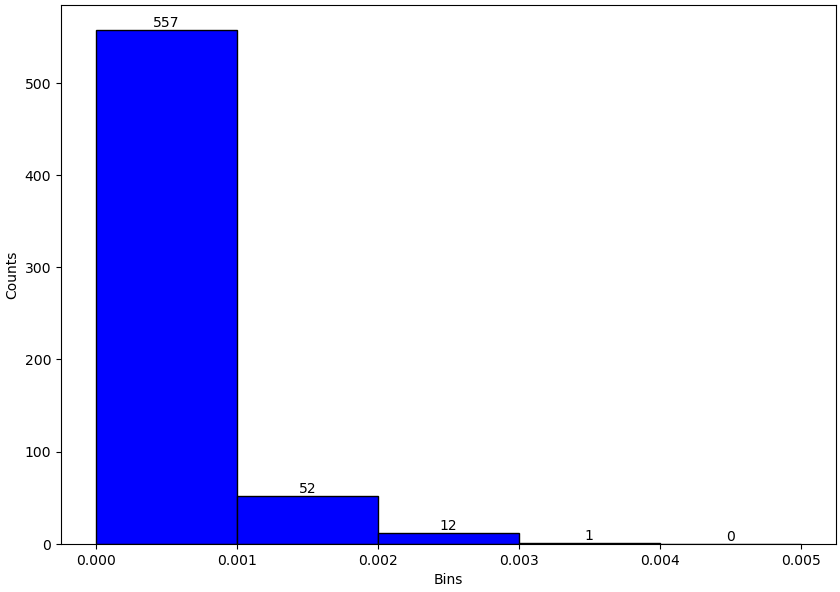} \\
\includegraphics[width=0.43\textwidth]{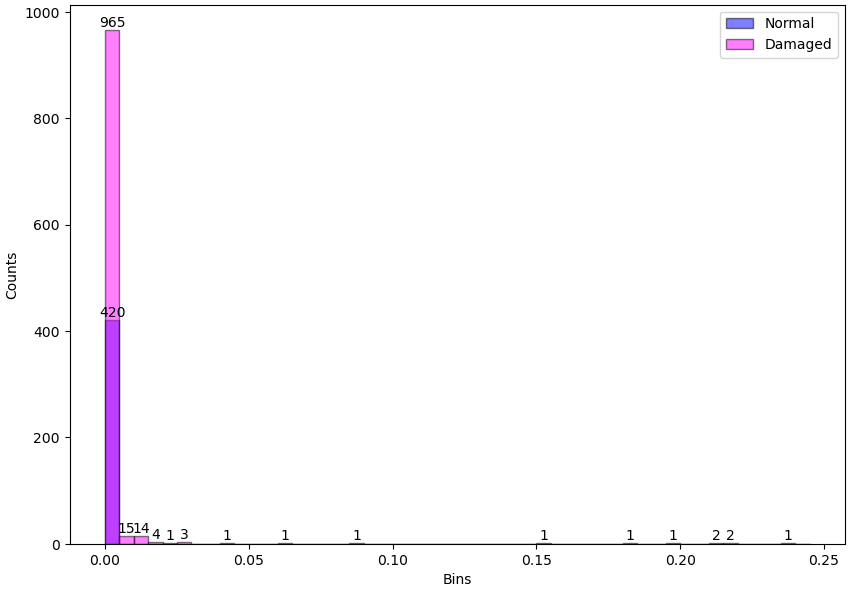}\\
\end{tabular}
\vspace{-6.9mm}
\caption{The top panel shows the histogram of reconstruction errors over the training dataset for an AE trained over this dataset. The bottom panel shows the histogram of AE reconstruction errors over the testing dataset. The reconstruction errors of damaged samples are superimposed over those of normal samples in the testing dataset.}
\label{fig:BridgeMSEHistograms}
\end{figure}

\begin{figure}
\centering
\begin{tabular}{cc}
\includegraphics[width=0.43\textwidth]{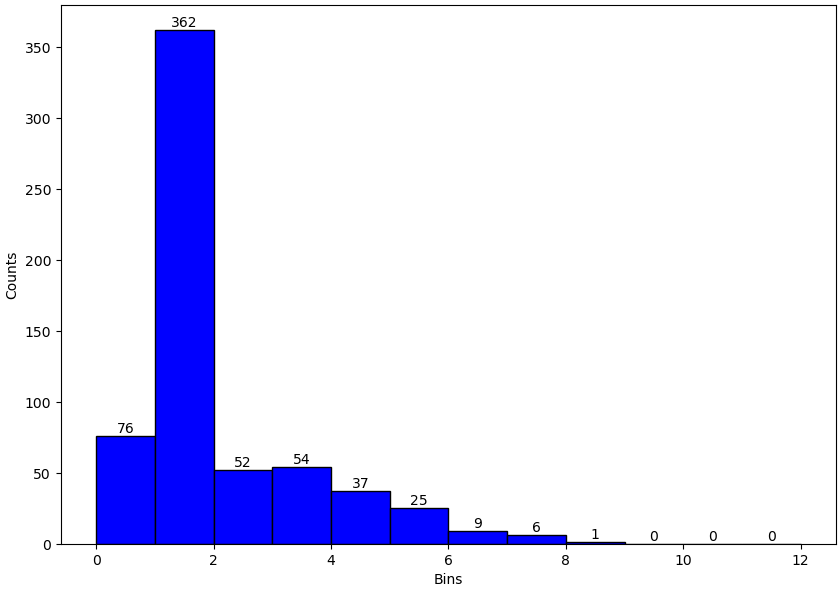} \\
\includegraphics[width=0.43\textwidth]{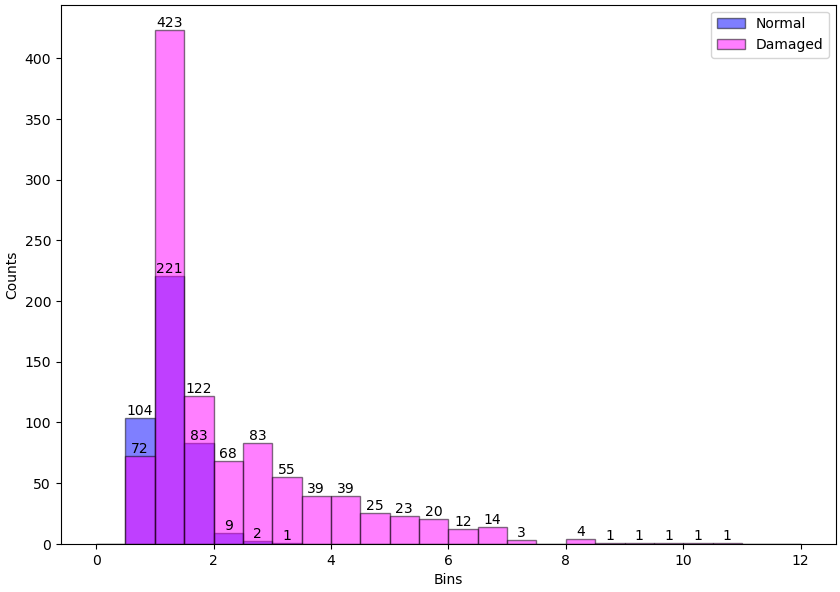}\\
\end{tabular}
\vspace*{-6.9mm}
\caption{The top panel shows the histogram of LOFs over the training dataset for an AE trained over this dataset. The bottom panel shows the histogram of LOFs over the testing dataset. The LOFs of damaged samples are superimposed over those of normal samples in the testing dataset.}
\label{fig:BridgeLOFHistograms}
\end{figure}

In this dataset, there is a significant difference in performance between StrOUD and MIL. The lower AUC values obtained by StrOUD indicate that, although the measurements are labeled according to the bridge health status, they are likely not accurately assigned. In fact, our detailed analysis reveals that even though the bridge has been damaged, most of the anomalous vibrations resemble normal segments. We relate here one dimension of this analysis that supports this characterization. We train an AE on the normal instances of the bridge data, and then relate reconstruction errors of normal versus damaged samples. The top panel of Fig.~\ref{fig:BridgeMSEHistograms} shows the histogram of reconstruction errors over the training dataset. The bottom panel shows the histogram of reconstruction errors over the testing dataset. For the latter, the reconstruction errors of samples labeled as damaged are superimposed over those of samples labeled as normal.  Fig.~\ref{fig:BridgeLOFHistograms} relates a similar analysis but on the LOFs. 

Fig.~\ref{fig:BridgeMSEHistograms}-\ref{fig:BridgeLOFHistograms} clearly show that it is hard to distinguish normal from damaged/anomalous samples. In fact, this analysis additionally makes the case why threshold-based AD methods, whether they utilize LOF or AE reconstruction error are not effective for real-world data. This particular dataset demonstrates that the location, spatial orientation, and proximity of a sensor to the defect will lead to very different signal response. Additionally, a sensor will not continuously respond the same way to a defect.

It is very clear from the Bridge dataset that, when we are dealing with raw real-world or manually-labeled datasets, evaluation becomes a challenging task due to very complex physical responses, and sometimes the uncertainty about label accuracy. However, such an issue is alleviated under the MIL formulation, as MIL depends on group rather than single-instance labels.

\section{Conclusion}
\label{sec:Conclusion}

Though largely under-leveraged for AD, we present in this paper an MIL formulation for AD and pursue an empirical evaluation of various algorithmic instantiations under this formulation. We evaluate the resulting algorithms over four different datasets that capture different physical processes along different modalities. The results show the superiority of the MIL-based formulation over single instance learning, as well as the ability of the MIL framework to generalize well over diverse benchmark datasets resulting from different real-world application domains. While in this paper we have focused the comparison against StrOUD, effectively carrying out an ablation study, future work will consider expanding comparisons to other AD algorithms to obtain a better understanding of their relationship with real-world datasets with varying label accuracy confidence. Future work will also leverage the versatility of the MIL framework and consider different strangeness score formulations. 

\bibliographystyle{ACM-Reference-Format}
\bibliography{sample-base}


\begin{thebibliography}{43}


\ifx \showCODEN    \undefined \def \showCODEN     #1{\unskip}     \fi
\ifx \showDOI      \undefined \def \showDOI       #1{#1}\fi
\ifx \showISBNx    \undefined \def \showISBNx     #1{\unskip}     \fi
\ifx \showISBNxiii \undefined \def \showISBNxiii  #1{\unskip}     \fi
\ifx \showISSN     \undefined \def \showISSN      #1{\unskip}     \fi
\ifx \showLCCN     \undefined \def \showLCCN      #1{\unskip}     \fi
\ifx \shownote     \undefined \def \shownote      #1{#1}          \fi
\ifx \showarticletitle \undefined \def \showarticletitle #1{#1}   \fi
\ifx \showURL      \undefined \def \showURL       {\relax}        \fi
\providecommand\bibfield[2]{#2}
\providecommand\bibinfo[2]{#2}
\providecommand\natexlab[1]{#1}
\providecommand\showeprint[2][]{arXiv:#2}

\bibitem[Ahmed et~al\mbox{.}(2017)]%
        {ahmed2017anomaly}
\bibfield{author}{\bibinfo{person}{Mohiuddin Ahmed}, \bibinfo{person}{Nazim
  Choudhury}, {and} \bibinfo{person}{Shahadat Uddin}.}
  \bibinfo{year}{2017}\natexlab{}.
\newblock \showarticletitle{Anomaly detection on big data in financial
  markets}. In \bibinfo{booktitle}{\emph{2017 IEEE/ACM International Conference
  on Advances in Social Networks Analysis and Mining (ASONAM)}}. IEEE,
  \bibinfo{pages}{998--1001}.
\newblock


\bibitem[Alam and Shehu(2020)]%
        {AlamShehuBICOB20}
\bibfield{author}{\bibinfo{person}{F.~F. Alam} {and} \bibinfo{person}{A.
  Shehu}.} \bibinfo{year}{2020}\natexlab{}.
\newblock \showarticletitle{From Unsupervised Multi-Instance Learning to
  Identification of Near-Native Protein Structures}. In
  \bibinfo{booktitle}{\emph{Intl Conf on Bioinf and Comput Biol}},
  Vol.~\bibinfo{volume}{70}. \bibinfo{pages}{59--68}.
\newblock


\bibitem[Alam and Shehu(2021)]%
        {AlamShehuJBCB21}
\bibfield{author}{\bibinfo{person}{F.~F. Alam} {and} \bibinfo{person}{A.
  Shehu}.} \bibinfo{year}{2021}\natexlab{}.
\newblock \showarticletitle{Unsupervised Multi-Instance Learning for Protein
  Structure Determination}.
\newblock \bibinfo{journal}{\emph{J Bioinf \& Comput Biol}}
  \bibinfo{volume}{19}, \bibinfo{number}{1} (\bibinfo{year}{2021}),
  \bibinfo{pages}{2140002}.
\newblock


\bibitem[Amores(2013)]%
        {amores2013multiple}
\bibfield{author}{\bibinfo{person}{Jaume Amores}.}
  \bibinfo{year}{2013}\natexlab{}.
\newblock \showarticletitle{Multiple instance classification: Review, taxonomy
  and comparative study}.
\newblock \bibinfo{journal}{\emph{Artificial intelligence}}
  \bibinfo{volume}{201} (\bibinfo{year}{2013}), \bibinfo{pages}{81--105}.
\newblock


\bibitem[An and Cho(2015)]%
        {an2015variational}
\bibfield{author}{\bibinfo{person}{Jinwon An} {and} \bibinfo{person}{Sungzoon
  Cho}.} \bibinfo{year}{2015}\natexlab{}.
\newblock \showarticletitle{Variational autoencoder based anomaly detection
  using reconstruction probability}.
\newblock \bibinfo{journal}{\emph{Special Lecture on IE}} \bibinfo{volume}{2},
  \bibinfo{number}{1} (\bibinfo{year}{2015}), \bibinfo{pages}{1--18}.
\newblock


\bibitem[Andrews et~al\mbox{.}(2002)]%
        {andrews2002support}
\bibfield{author}{\bibinfo{person}{Stuart Andrews}, \bibinfo{person}{Ioannis
  Tsochantaridis}, {and} \bibinfo{person}{Thomas Hofmann}.}
  \bibinfo{year}{2002}\natexlab{}.
\newblock \showarticletitle{Support vector machines for multiple-instance
  learning}.
\newblock \bibinfo{journal}{\emph{Advances in neural information processing
  systems}}  \bibinfo{volume}{15} (\bibinfo{year}{2002}).
\newblock


\bibitem[Babaei et~al\mbox{.}(2021)]%
        {babaei2021aegr}
\bibfield{author}{\bibinfo{person}{Kasra Babaei}, \bibinfo{person}{Zhi~Yuan
  Chen}, {and} \bibinfo{person}{Tomas Maul}.} \bibinfo{year}{2021}\natexlab{}.
\newblock \showarticletitle{AEGR: A simple approach to gradient reversal in
  autoencoders for network anomaly detection}.
\newblock \bibinfo{journal}{\emph{Soft Computing}} \bibinfo{volume}{25},
  \bibinfo{number}{24} (\bibinfo{year}{2021}), \bibinfo{pages}{15269--15280}.
\newblock


\bibitem[Barbar{\'a} et~al\mbox{.}(2006)]%
        {barbara2006detecting}
\bibfield{author}{\bibinfo{person}{Daniel Barbar{\'a}},
  \bibinfo{person}{Carlotta Domeniconi}, {and} \bibinfo{person}{James~P
  Rogers}.} \bibinfo{year}{2006}\natexlab{}.
\newblock \showarticletitle{Detecting outliers using transduction and
  statistical testing}. In \bibinfo{booktitle}{\emph{Proceedings of the 12th
  ACM SIGKDD international conference on Knowledge discovery and data mining}}.
  \bibinfo{pages}{55--64}.
\newblock


\bibitem[Bl{\'a}zquez-Garc{\'\i}a et~al\mbox{.}(2021)]%
        {blazquez2021review}
\bibfield{author}{\bibinfo{person}{Ane Bl{\'a}zquez-Garc{\'\i}a},
  \bibinfo{person}{Angel Conde}, \bibinfo{person}{Usue Mori}, {and}
  \bibinfo{person}{Jose~A Lozano}.} \bibinfo{year}{2021}\natexlab{}.
\newblock \showarticletitle{A review on outlier/anomaly detection in time
  series data}.
\newblock \bibinfo{journal}{\emph{ACM Computing Surveys (CSUR)}}
  \bibinfo{volume}{54}, \bibinfo{number}{3} (\bibinfo{year}{2021}),
  \bibinfo{pages}{1--33}.
\newblock


\bibitem[Breunig et~al\mbox{.}(2000)]%
        {breunig2000lof}
\bibfield{author}{\bibinfo{person}{Markus~M Breunig},
  \bibinfo{person}{Hans-Peter Kriegel}, \bibinfo{person}{Raymond~T Ng}, {and}
  \bibinfo{person}{J{\"o}rg Sander}.} \bibinfo{year}{2000}\natexlab{}.
\newblock \showarticletitle{LOF: identifying density-based local outliers}. In
  \bibinfo{booktitle}{\emph{Proceedings of the 2000 ACM SIGMOD international
  conference on Management of data}}. \bibinfo{pages}{93--104}.
\newblock


\bibitem[Carbonneau et~al\mbox{.}(2018)]%
        {carbonneau2018multiple}
\bibfield{author}{\bibinfo{person}{Marc-Andr{\'e} Carbonneau},
  \bibinfo{person}{Veronika Cheplygina}, \bibinfo{person}{Eric Granger}, {and}
  \bibinfo{person}{Ghyslain Gagnon}.} \bibinfo{year}{2018}\natexlab{}.
\newblock \showarticletitle{Multiple instance learning: A survey of problem
  characteristics and applications}.
\newblock \bibinfo{journal}{\emph{Pattern Recognition}}  \bibinfo{volume}{77}
  (\bibinfo{year}{2018}), \bibinfo{pages}{329--353}.
\newblock


\bibitem[Chandola et~al\mbox{.}(2009a)]%
        {chandola2009anomaly}
\bibfield{author}{\bibinfo{person}{Varun Chandola}, \bibinfo{person}{Arindam
  Banerjee}, {and} \bibinfo{person}{Vipin Kumar}.}
  \bibinfo{year}{2009}\natexlab{a}.
\newblock \showarticletitle{Anomaly detection: A survey}.
\newblock \bibinfo{journal}{\emph{ACM computing surveys (CSUR)}}
  \bibinfo{volume}{41}, \bibinfo{number}{3} (\bibinfo{year}{2009}),
  \bibinfo{pages}{1--58}.
\newblock


\bibitem[Chandola et~al\mbox{.}(2009b)]%
        {chandola2009detecting}
\bibfield{author}{\bibinfo{person}{Varun Chandola}, \bibinfo{person}{Deepthi
  Cheboli}, {and} \bibinfo{person}{Vipin Kumar}.}
  \bibinfo{year}{2009}\natexlab{b}.
\newblock \showarticletitle{Detecting anomalies in a time series database}.
\newblock  (\bibinfo{year}{2009}).
\newblock


\bibitem[Chandola et~al\mbox{.}(2008)]%
        {chandola2008comparative}
\bibfield{author}{\bibinfo{person}{Varun Chandola}, \bibinfo{person}{Varun
  Mithal}, {and} \bibinfo{person}{Vipin Kumar}.}
  \bibinfo{year}{2008}\natexlab{}.
\newblock \showarticletitle{Comparative evaluation of anomaly detection
  techniques for sequence data}. In \bibinfo{booktitle}{\emph{2008 Eighth IEEE
  international conference on data mining}}. IEEE, \bibinfo{pages}{743--748}.
\newblock


\bibitem[Chen et~al\mbox{.}(2018)]%
        {chen2018autoencoder}
\bibfield{author}{\bibinfo{person}{Zhaomin Chen}, \bibinfo{person}{Chai~Kiat
  Yeo}, \bibinfo{person}{Bu~Sung Lee}, {and} \bibinfo{person}{Chiew~Tong Lau}.}
  \bibinfo{year}{2018}\natexlab{}.
\newblock \showarticletitle{Autoencoder-based network anomaly detection}. In
  \bibinfo{booktitle}{\emph{2018 Wireless Telecommunications Symposium (WTS)}}.
  IEEE, \bibinfo{pages}{1--5}.
\newblock


\bibitem[da~Costa et~al\mbox{.}(2020)]%
        {da2020critical}
\bibfield{author}{\bibinfo{person}{Kelton~AP da Costa},
  \bibinfo{person}{Jo{\~a}o~P Papa}, \bibinfo{person}{Leandro~A Passos},
  \bibinfo{person}{Danilo Colombo}, \bibinfo{person}{Javier Del~Ser},
  \bibinfo{person}{Khan Muhammad}, {and} \bibinfo{person}{Victor Hugo~C de
  Albuquerque}.} \bibinfo{year}{2020}\natexlab{}.
\newblock \showarticletitle{A critical literature survey and prospects on
  tampering and anomaly detection in image data}.
\newblock \bibinfo{journal}{\emph{Applied Soft Computing}}
  \bibinfo{volume}{97} (\bibinfo{year}{2020}), \bibinfo{pages}{106727}.
\newblock


\bibitem[Dietterich et~al\mbox{.}(1997)]%
        {dietterich1997solving}
\bibfield{author}{\bibinfo{person}{Thomas~G Dietterich},
  \bibinfo{person}{Richard~H Lathrop}, {and} \bibinfo{person}{Tom{\'a}s
  Lozano-P{\'e}rez}.} \bibinfo{year}{1997}\natexlab{}.
\newblock \showarticletitle{Solving the multiple instance problem with
  axis-parallel rectangles}.
\newblock \bibinfo{journal}{\emph{Artificial intelligence}}
  \bibinfo{volume}{89}, \bibinfo{number}{1-2} (\bibinfo{year}{1997}),
  \bibinfo{pages}{31--71}.
\newblock


\bibitem[Du~Prel et~al\mbox{.}(2009)]%
        {du2009confidence}
\bibfield{author}{\bibinfo{person}{Jean-Baptist Du~Prel},
  \bibinfo{person}{Gerhard Hommel}, \bibinfo{person}{Bernd R{\"o}hrig}, {and}
  \bibinfo{person}{Maria Blettner}.} \bibinfo{year}{2009}\natexlab{}.
\newblock \showarticletitle{Confidence interval or p-value?: part 4 of a series
  on evaluation of scientific publications}.
\newblock \bibinfo{journal}{\emph{Deutsches {\"A}rzteblatt International}}
  \bibinfo{volume}{106}, \bibinfo{number}{19} (\bibinfo{year}{2009}),
  \bibinfo{pages}{335}.
\newblock


\bibitem[Dua and Graff(2017)]%
        {Dua:2019}
\bibfield{author}{\bibinfo{person}{Dheeru Dua} {and} \bibinfo{person}{Casey
  Graff}.} \bibinfo{year}{2017}\natexlab{}.
\newblock \bibinfo{title}{{UCI} Machine Learning Repository}.
\newblock
\newblock
\urldef\tempurl%
\url{http://archive.ics.uci.edu/ml}
\showURL{%
\tempurl}


\bibitem[Feng et~al\mbox{.}(2021a)]%
        {feng2021mist}
\bibfield{author}{\bibinfo{person}{Jia-Chang Feng}, \bibinfo{person}{Fa-Ting
  Hong}, {and} \bibinfo{person}{Wei-Shi Zheng}.}
  \bibinfo{year}{2021}\natexlab{a}.
\newblock \showarticletitle{Mist: Multiple instance self-training framework for
  video anomaly detection}. In \bibinfo{booktitle}{\emph{Proceedings of the
  IEEE/CVF conference on computer vision and pattern recognition}}.
  \bibinfo{pages}{14009--14018}.
\newblock


\bibitem[Feng et~al\mbox{.}(2021b)]%
        {feng2021multiple}
\bibfield{author}{\bibinfo{person}{Lei Feng}, \bibinfo{person}{Senlin Shu},
  \bibinfo{person}{Yuzhou Cao}, \bibinfo{person}{Lue Tao},
  \bibinfo{person}{Hongxin Wei}, \bibinfo{person}{Tao Xiang},
  \bibinfo{person}{Bo An}, {and} \bibinfo{person}{Gang Niu}.}
  \bibinfo{year}{2021}\natexlab{b}.
\newblock \showarticletitle{Multiple-Instance Learning from Similar and
  Dissimilar Bags}. In \bibinfo{booktitle}{\emph{Proceedings of the 27th ACM
  SIGKDD Conference on Knowledge Discovery \& Data Mining}}.
  \bibinfo{pages}{374--382}.
\newblock


\bibitem[Foulds and Frank(2010)]%
        {foulds2010review}
\bibfield{author}{\bibinfo{person}{James Foulds} {and} \bibinfo{person}{Eibe
  Frank}.} \bibinfo{year}{2010}\natexlab{}.
\newblock \showarticletitle{A review of multi-instance learning assumptions}.
\newblock \bibinfo{journal}{\emph{The knowledge engineering review}}
  \bibinfo{volume}{25}, \bibinfo{number}{1} (\bibinfo{year}{2010}),
  \bibinfo{pages}{1--25}.
\newblock


\bibitem[Guan et~al\mbox{.}(2016)]%
        {guan2016efficient}
\bibfield{author}{\bibinfo{person}{Xinze Guan}, \bibinfo{person}{Raviv Raich},
  {and} \bibinfo{person}{Weng-Keen Wong}.} \bibinfo{year}{2016}\natexlab{}.
\newblock \showarticletitle{Efficient multi-instance learning for activity
  recognition from time series data using an auto-regressive hidden markov
  model}. In \bibinfo{booktitle}{\emph{International Conference on Machine
  Learning}}. PMLR, \bibinfo{pages}{2330--2339}.
\newblock


\bibitem[Hautamaki et~al\mbox{.}(2004)]%
        {hautamaki2004outlier}
\bibfield{author}{\bibinfo{person}{Ville Hautamaki}, \bibinfo{person}{Ismo
  Karkkainen}, {and} \bibinfo{person}{Pasi Franti}.}
  \bibinfo{year}{2004}\natexlab{}.
\newblock \showarticletitle{Outlier detection using k-nearest neighbour graph}.
  In \bibinfo{booktitle}{\emph{Proceedings of the 17th International Conference
  on Pattern Recognition, 2004. ICPR 2004.}}, Vol.~\bibinfo{volume}{3}. IEEE,
  \bibinfo{pages}{430--433}.
\newblock


\bibitem[Id{\'e} et~al\mbox{.}(2007)]%
        {ide2007computing}
\bibfield{author}{\bibinfo{person}{Tsuyoshi Id{\'e}}, \bibinfo{person}{Spiros
  Papadimitriou}, {and} \bibinfo{person}{Michail Vlachos}.}
  \bibinfo{year}{2007}\natexlab{}.
\newblock \showarticletitle{Computing correlation anomaly scores using
  stochastic nearest neighbors}. In \bibinfo{booktitle}{\emph{Seventh IEEE
  international conference on data mining (ICDM 2007)}}. IEEE,
  \bibinfo{pages}{523--528}.
\newblock


\bibitem[Kim et~al\mbox{.}(2021)]%
        {kim2021ambient}
\bibfield{author}{\bibinfo{person}{Chul-Woo Kim}, \bibinfo{person}{Feng-Liang
  Zhang}, \bibinfo{person}{Kai-Chun Chang}, \bibinfo{person}{Patrick~John
  McGetrick}, {and} \bibinfo{person}{Yoshinao Goi}.}
  \bibinfo{year}{2021}\natexlab{}.
\newblock \showarticletitle{Ambient and vehicle-induced vibration data of a
  steel truss bridge subject to artificial damage}.
\newblock \bibinfo{journal}{\emph{Journal of Bridge Engineering}}
  \bibinfo{volume}{26}, \bibinfo{number}{7} (\bibinfo{year}{2021}),
  \bibinfo{pages}{04721002}.
\newblock


\bibitem[Kou et~al\mbox{.}(2004)]%
        {kou2004survey}
\bibfield{author}{\bibinfo{person}{Yufeng Kou}, \bibinfo{person}{Chang-Tien
  Lu}, \bibinfo{person}{Sirirat Sirwongwattana}, {and} \bibinfo{person}{Yo-Ping
  Huang}.} \bibinfo{year}{2004}\natexlab{}.
\newblock \showarticletitle{Survey of fraud detection techniques}. In
  \bibinfo{booktitle}{\emph{IEEE international conference on networking,
  sensing and control, 2004}}, Vol.~\bibinfo{volume}{2}. IEEE,
  \bibinfo{pages}{749--754}.
\newblock


\bibitem[Laxhammar and Falkman(2011)]%
        {laxhammar2011sequential}
\bibfield{author}{\bibinfo{person}{Rikard Laxhammar} {and}
  \bibinfo{person}{G{\"o}ran Falkman}.} \bibinfo{year}{2011}\natexlab{}.
\newblock \showarticletitle{Sequential conformal anomaly detection in
  trajectories based on hausdorff distance}. In \bibinfo{booktitle}{\emph{14th
  International Conference on Information Fusion}}. IEEE,
  \bibinfo{pages}{1--8}.
\newblock


\bibitem[Laxhammar and Falkman(2015)]%
        {laxhammar2015inductive}
\bibfield{author}{\bibinfo{person}{Rikard Laxhammar} {and}
  \bibinfo{person}{G{\"o}ran Falkman}.} \bibinfo{year}{2015}\natexlab{}.
\newblock \showarticletitle{Inductive conformal anomaly detection for
  sequential detection of anomalous sub-trajectories}.
\newblock \bibinfo{journal}{\emph{Annals of Mathematics and Artificial
  Intelligence}} \bibinfo{volume}{74}, \bibinfo{number}{1}
  (\bibinfo{year}{2015}), \bibinfo{pages}{67--94}.
\newblock


\bibitem[Oh et~al\mbox{.}(2011)]%
        {oh2011large}
\bibfield{author}{\bibinfo{person}{Sangmin Oh}, \bibinfo{person}{Anthony
  Hoogs}, \bibinfo{person}{Amitha Perera}, \bibinfo{person}{Naresh Cuntoor},
  \bibinfo{person}{Chia-Chih Chen}, \bibinfo{person}{Jong~Taek Lee},
  \bibinfo{person}{Saurajit Mukherjee}, \bibinfo{person}{JK Aggarwal},
  \bibinfo{person}{Hyungtae Lee}, \bibinfo{person}{Larry Davis},
  {et~al\mbox{.}}} \bibinfo{year}{2011}\natexlab{}.
\newblock \showarticletitle{A large-scale benchmark dataset for event
  recognition in surveillance video}. In \bibinfo{booktitle}{\emph{CVPR 2011}}.
  IEEE, \bibinfo{pages}{3153--3160}.
\newblock


\bibitem[Quellec et~al\mbox{.}(2016)]%
        {quellec2016multiple}
\bibfield{author}{\bibinfo{person}{Gwenol{\'e} Quellec},
  \bibinfo{person}{Mathieu Lamard}, \bibinfo{person}{Michel Cozic},
  \bibinfo{person}{Gouenou Coatrieux}, {and} \bibinfo{person}{Guy Cazuguel}.}
  \bibinfo{year}{2016}\natexlab{}.
\newblock \showarticletitle{Multiple-instance learning for anomaly detection in
  digital mammography}.
\newblock \bibinfo{journal}{\emph{Ieee transactions on medical imaging}}
  \bibinfo{volume}{35}, \bibinfo{number}{7} (\bibinfo{year}{2016}),
  \bibinfo{pages}{1604--1614}.
\newblock


\bibitem[Shyu et~al\mbox{.}(2003)]%
        {shyu2003novel}
\bibfield{author}{\bibinfo{person}{Mei-Ling Shyu}, \bibinfo{person}{Shu-Ching
  Chen}, \bibinfo{person}{Kanoksri Sarinnapakorn}, {and} \bibinfo{person}{LiWu
  Chang}.} \bibinfo{year}{2003}\natexlab{}.
\newblock \bibinfo{booktitle}{\emph{A novel anomaly detection scheme based on
  principal component classifier}}.
\newblock \bibinfo{type}{{T}echnical {R}eport}. \bibinfo{institution}{Miami
  Univ Coral Gables Fl Dept of Electrical and Computer Engineering}.
\newblock


\bibitem[Sultani et~al\mbox{.}(2018)]%
        {sultani2018real}
\bibfield{author}{\bibinfo{person}{Waqas Sultani}, \bibinfo{person}{Chen Chen},
  {and} \bibinfo{person}{Mubarak Shah}.} \bibinfo{year}{2018}\natexlab{}.
\newblock \showarticletitle{Real-world anomaly detection in surveillance
  videos}. In \bibinfo{booktitle}{\emph{Proceedings of the IEEE conference on
  computer vision and pattern recognition}}. \bibinfo{pages}{6479--6488}.
\newblock


\bibitem[Szegedy et~al\mbox{.}(2016)]%
        {szegedy2016rethinking}
\bibfield{author}{\bibinfo{person}{Christian Szegedy}, \bibinfo{person}{Vincent
  Vanhoucke}, \bibinfo{person}{Sergey Ioffe}, \bibinfo{person}{Jon Shlens},
  {and} \bibinfo{person}{Zbigniew Wojna}.} \bibinfo{year}{2016}\natexlab{}.
\newblock \showarticletitle{Rethinking the inception architecture for computer
  vision}. In \bibinfo{booktitle}{\emph{Proceedings of the IEEE conference on
  computer vision and pattern recognition}}. \bibinfo{pages}{2818--2826}.
\newblock


\bibitem[Tang et~al\mbox{.}(2002)]%
        {tang2002enhancing}
\bibfield{author}{\bibinfo{person}{Jian Tang}, \bibinfo{person}{Zhixiang Chen},
  \bibinfo{person}{Ada Wai-Chee Fu}, {and} \bibinfo{person}{David~W Cheung}.}
  \bibinfo{year}{2002}\natexlab{}.
\newblock \showarticletitle{Enhancing effectiveness of outlier detections for
  low density patterns}. In \bibinfo{booktitle}{\emph{Pacific-Asia Conference
  on Knowledge Discovery and Data Mining}}. Springer,
  \bibinfo{pages}{535--548}.
\newblock


\bibitem[Tschannen et~al\mbox{.}(2018)]%
        {tschannen2018recent}
\bibfield{author}{\bibinfo{person}{Michael Tschannen}, \bibinfo{person}{Olivier
  Bachem}, {and} \bibinfo{person}{Mario Lucic}.}
  \bibinfo{year}{2018}\natexlab{}.
\newblock \showarticletitle{Recent advances in autoencoder-based representation
  learning}.
\newblock \bibinfo{journal}{\emph{arXiv preprint arXiv:1812.05069}}
  (\bibinfo{year}{2018}).
\newblock


\bibitem[Villa-P{\'e}rez et~al\mbox{.}(2021)]%
        {villa2021semi}
\bibfield{author}{\bibinfo{person}{Miryam~Elizabeth Villa-P{\'e}rez},
  \bibinfo{person}{Miguel~{\'A} {\'A}lvarez-Carmona}, \bibinfo{person}{Octavio
  Loyola-Gonz{\'a}lez}, \bibinfo{person}{Miguel~Angel Medina-P{\'e}rez},
  \bibinfo{person}{Juan~Carlos Velazco-Rossell}, {and}
  \bibinfo{person}{Kim-Kwang~Raymond Choo}.} \bibinfo{year}{2021}\natexlab{}.
\newblock \showarticletitle{Semi-supervised anomaly detection algorithms: A
  comparative summary and future research directions}.
\newblock \bibinfo{journal}{\emph{Knowledge-Based Systems}}
  \bibinfo{volume}{218} (\bibinfo{year}{2021}), \bibinfo{pages}{106878}.
\newblock


\bibitem[Yang et~al\mbox{.}(2013)]%
        {yang2013TRASMIL}
\bibfield{author}{\bibinfo{person}{Wanki Yang}, \bibinfo{person}{Yang Gao},
  {and} \bibinfo{person}{Longbing Cao}.} \bibinfo{year}{2013}\natexlab{}.
\newblock \showarticletitle{TRASMIL: A local anomaly detection framework based
  on trajectory segmentation and multi-instance learning}.
\newblock \bibinfo{journal}{\emph{Computer Vision and Image Understanding}}
  \bibinfo{volume}{117}, \bibinfo{number}{10} (\bibinfo{year}{2013}),
  \bibinfo{pages}{1273--1286}.
\newblock


\bibitem[Yang et~al\mbox{.}(2018)]%
        {yang2018bearing}
\bibfield{author}{\bibinfo{person}{Yanli Yang}, \bibinfo{person}{Peiying Fu},
  {and} \bibinfo{person}{Yichuan He}.} \bibinfo{year}{2018}\natexlab{}.
\newblock \showarticletitle{Bearing fault automatic classification based on
  deep learning}.
\newblock \bibinfo{journal}{\emph{IEEE Access}}  \bibinfo{volume}{6}
  (\bibinfo{year}{2018}), \bibinfo{pages}{71540--71554}.
\newblock


\bibitem[Zenati et~al\mbox{.}(2018)]%
        {zenati2018adversarially}
\bibfield{author}{\bibinfo{person}{Houssam Zenati}, \bibinfo{person}{Manon
  Romain}, \bibinfo{person}{Chuan-Sheng Foo}, \bibinfo{person}{Bruno Lecouat},
  {and} \bibinfo{person}{Vijay Chandrasekhar}.}
  \bibinfo{year}{2018}\natexlab{}.
\newblock \showarticletitle{Adversarially learned anomaly detection}. In
  \bibinfo{booktitle}{\emph{2018 IEEE International conference on data mining
  (ICDM)}}. IEEE, \bibinfo{pages}{727--736}.
\newblock


\bibitem[Zhang and Viola(2007)]%
        {zhang2007multiple}
\bibfield{author}{\bibinfo{person}{Cha Zhang} {and} \bibinfo{person}{Paul
  Viola}.} \bibinfo{year}{2007}\natexlab{}.
\newblock \showarticletitle{Multiple-instance pruning for learning efficient
  cascade detectors}.
\newblock \bibinfo{journal}{\emph{Advances in neural information processing
  systems}}  \bibinfo{volume}{20} (\bibinfo{year}{2007}).
\newblock


\bibitem[Zhang et~al\mbox{.}(2021b)]%
        {zhang2021multiple}
\bibfield{author}{\bibinfo{person}{Jianyi Zhang}, \bibinfo{person}{Jiaqi Yao},
  \bibinfo{person}{Yanjie Chu}, {and} \bibinfo{person}{Jikun Yan}.}
  \bibinfo{year}{2021}\natexlab{b}.
\newblock \showarticletitle{A Multiple Instance Learning Algorithm Using Graph
  Convolutional Network for Speech Content Classification}. In
  \bibinfo{booktitle}{\emph{2021 IEEE 5th Information Technology, Networking,
  Electronic and Automation Control Conference (ITNEC)}},
  Vol.~\bibinfo{volume}{5}. IEEE, \bibinfo{pages}{1480--1484}.
\newblock


\bibitem[Zhang et~al\mbox{.}(2021a)]%
        {zhang2021anomaly}
\bibfield{author}{\bibinfo{person}{Yi-Ming Zhang}, \bibinfo{person}{Hao Wang},
  \bibinfo{person}{Hua-Ping Wan}, \bibinfo{person}{Jian-Xiao Mao}, {and}
  \bibinfo{person}{Yi-Chao Xu}.} \bibinfo{year}{2021}\natexlab{a}.
\newblock \showarticletitle{Anomaly detection of structural health monitoring
  data using the maximum likelihood estimation-based Bayesian dynamic linear
  model}.
\newblock \bibinfo{journal}{\emph{Structural Health Monitoring}}
  \bibinfo{volume}{20}, \bibinfo{number}{6} (\bibinfo{year}{2021}),
  \bibinfo{pages}{2936--2952}.
\newblock


\end{thebibliography}

\end{document}